\documentclass[letterpaper, 10 pt, conference]{ieeeconf}  

\usepackage{fontawesome}
\def\pscircled#1{\textcircled{\resizebox{.5em}{!}{#1}}}

\usepackage{cite}  
\IEEEoverridecommandlockouts                              
\usepackage{graphics} 
\usepackage{epsfig}   
\usepackage{mathptmx} 
\usepackage{times}    
\usepackage{amsmath}  
\usepackage{amssymb}  
\usepackage[utf8]{inputenc}
\usepackage[english]{babel}
\usepackage{float}
\usepackage{amsfonts}
\usepackage{booktabs}
\usepackage{epsfig}
\usepackage{epstopdf}
\usepackage{subfigure}
\usepackage{cite}
\usepackage{multirow} 
\usepackage{algorithm,algpseudocode}
\usepackage{subcaption}
\usepackage{xcolor}

\usepackage[linkcolor=blue,colorlinks=true]{hyperref}

\newtheorem{definition}{Definition}
\newtheorem{theorem}{Theorem}
\newtheorem{lemma}[theorem]{Lemma}

\newtheorem{proposition}{Proposition}[definition]

\usepackage{caption}
\title{\LARGE \bf
More Optimal Fractional-Order Stochastic Gradient Descent for Non-Convex Optimization Problems
}

\author{Mohammad Partohaghighi$^{1}$, Roummel Marcia$^{2}$, and YangQuan Chen$^{1,3}$
\thanks{*Corresponding author: YangQuan Chen.  \pscircled{\faPhone}~{\tt +1-209-2284672}.
YC is supported by the Center for Methane Emission Research and Innovation (\href{http://methane.ucmerced.edu}{\tt CMERI}) through the Climate Action Seed Funds grant (2023-2026) at University of California, Merced. \pscircled{\faDesktop}~\href{http://mechatronics.ucmerced.edu}{\tt mechatronics.ucmerced.edu} }
\thanks{$^{1}$Mohammad Partohaghighi and YangQuan Chen are with the Electrical Engineering and Computer Science graduate program, 
        University of California at Merced, USA 
          \pscircled{\faEnvelope}~{\tt\small mpartohaghighi@ucmerced.edu}}%
\thanks{$^{2}$Roummel Marcia is with the Department of Applied Mathematics, 
        University of California Merced, Merced, CA, USA 
          \pscircled{\faEnvelope}~{\tt\small rmarcia@ucmerced.edu}}%
\thanks{$^{3}$YangQuan Chen is with the Mechatronics, Embedded Systems and Automation (MESA) Lab,         Department of Mechanical Engineering, School of Engineering, 
        University of California, Merced, CA 95343, USA 
          \pscircled{\faEnvelope}~{\tt\small ychen53@ucmerced.edu}}%
}

\begin{document}

\maketitle
\thispagestyle{empty}
\pagestyle{empty}
\begin{abstract}
Fractional-order stochastic gradient descent (FOSGD) leverages fractional exponents to capture long-memory effects in optimization. However, its utility is often limited by the difficulty of tuning and stabilizing these exponents. We propose 2SED Fractional-Order Stochastic Gradient Descent (2SEDFOSGD), which integrates the Two-Scale Effective Dimension (2SED) algorithm with FOSGD to adapt the fractional exponent in a data-driven manner. By tracking model sensitivity and effective dimensionality, 2SEDFOSGD dynamically modulates the exponent to mitigate oscillations and hasten convergence. Theoretically, for onoconvex optimization problems, this approach preserves the advantages of fractional memory without the sluggish or unstable behavior observed in naïve fractional SGD. Empirical evaluations in Gaussian and $\alpha$-stable noise scenarios using an autoregressive (AR) model highlight faster convergence and more robust parameter estimates compared to baseline methods, underscoring the potential of dimension-aware fractional techniques for advanced modeling and estimation tasks.
\end{abstract}


\renewcommand{\baselinestretch}{1}

\section{Introduction}

Machine learning (ML) and scientific computing increasingly rely on sophisticated optimization methods to tackle complex, high-dimensional problems. Classical stochastic gradient descent (SGD) has become a mainstay in training neural networks and large-scale models, owing to its simplicity and practical performance. However, standard SGD exhibits notable limitations: it typically treats updates as short-term corrections, discarding a rich history of past gradients. In contrast, fractional approaches in optimization draw upon the theory of fractional calculus to capture long-memory effects, thereby influencing the trajectory of updates by retaining historical gradient information over extended intervals \cite{zhou2025fractional, sheng2020fractional}.

Fractional calculus extends traditional calculus to include non-integer orders, offering a powerful tool for modeling and control in various fields, including optimization. It allows for the incorporation of memory and hereditary properties into models, which is particularly beneficial in dynamic systems and control applications \cite{Sheng2012, yin2014fractional}. By leveraging fractional derivatives, optimization algorithms can potentially achieve better convergence properties and robustness against noise, as they account for the accumulated effect of past gradients rather than relying solely on the most recent updates \cite{chen2018fractional2, liu2021quasi}. This approach has shown promise in enhancing the performance of optimization algorithms in machine learning and other scientific computing applications \cite{chen2017study}.

 By embracing these generalized derivatives, FOSGD modifies the usual gradient step to incorporate a partial summation of past gradients, effectively smoothing updates over a historical window. The method stands especially valuable for scenarios where prior states wield significant impact on the current gradient, as often encountered in dynamic processes or highly non-convex landscape \cite{shin2023accelerating}.
Nonetheless, the quest for harnessing fractional updates is not without drawbacks. Incorporating fractional operators demands added hyperparameters (particularly the fractional exponent $\alpha$), which can prove sensitive or unstable to tune. Excessively low or high fractional orders may slow convergence or lead to oscillatory gradients, thus negating the presumed benefits. Bridging the gap between the theoretical elegance of fractional calculus and the pressing computational demands of real-world ML systems remains a formidable challenge.

Although FOSGD helps mitigate short-term memory loss by preserving traces of past gradients, selecting and calibrating the fractional exponents can introduce substantial complexities in real-world settings. For instance, deciding whether \(\alpha = 0.5\) or \(\alpha = 0.9\) is most appropriate for capturing relevant memory structures is neither straightforward nor reliably robust, with the optimal choice often varying considerably across tasks or even across different stages of training. In practice, if the chosen exponent fails to align with the true dynamics of the loss landscape, updates may drift or stall, resulting in unpredictable or sluggish convergence. Moreover, fractional terms can amplify variance in gradient estimatesespecially under noisy or non-stationary conditions thereby causing oscillatory or chaotic training behaviors that undermine stability.

Beyond these convergence and stability concerns, fractional exponents impose additional burdens on tuning and hyperparameter selection. Even minor changes in \(\alpha\) can radically alter the memory effect, forcing practitioners to engage in extensive trial-and-error experiments to achieve consistent results. Such overhead becomes especially prohibitive in large-scale or time-sensitive applications, where iterating over a range of fractional parameters is not feasible. Consequently, despite its theoretical promise as a memory-based learning strategy, FOSGD faces limited adoption in practice, as the algorithm's strong reliance on well-chosen exponents can undercut the potential advantages that long-range gradient retention might otherwise provide.

A geometry-aware strategy like Two-Scale Effective Dimension (2SED) can dynamically regulate the fractional exponent in FOSGD. By examining partial diagonal approximations of the Fisher information matrix \cite{datres2024two}, 2SED identifies regions of high sensitivity and adapts the exponent accordingly. This approach dampens updates in areas prone to instability while exploiting longer memory in flatter regions. Consequently, combining 2SED with FOSGD reduces erratic oscillations, preserves long-term memory benefits, and yields more robust performance across diverse data sets and problem types.

We introduce a novel 2SED-driven FOSGD framework that dynamically regulates the fractional exponent using the dimension-aware metrics of 2SED. This adaptive mechanism aligns historical-gradient memory with the sensitivity of the optimization landscape, thereby enhancing stability and data alignment. Under standard smoothness and bounded-gradient assumptions, the method satisfies strong convergence criteria. In practice, geometry-based regularization fosters a more consistent convergence, as evidenced by solving an autoregressive (AR) model under  Gaussian and $\alpha$-stable noise.

We organize this paper as follows. In Section~\ref{sec:2sed_frac_sgd}, we thoroughly examine the 2SED algorithm, illustrating how it approximates second-order geometry to produce dimension-aware updates. Section~\ref{sec:frac_sgd_highlvl} reviews FOSGD, highlighting its appeal for long-memory processes and the hyperparameter dilemmas that hinder practicality. Also, we detail how to embed 2SED’s dimension metrics into the fractional framework, providing both equations and pseudo-code. Section~\ref{sec:convergence_convex} delves into a convergence analysis, establishing theoretical performance bounds for our method. Section~\ref{sec:example1} showcases experiments on synthetic and real-world tasks, confirming that 2SED-driven exponent adaptation yields measurably stronger results. Finally, Section~\ref{sec:conclusion} concludes by summarizing key findings, identifying broader implications for optimization, and suggesting directions for further research in advanced fractional calculus and dimension-based learning techniques.



\section{Two-Scale Effective Dimension (2SED) and FOSGD}
\label{sec:2sed_frac_sgd}

In modern deep learning, classical complexity measures such as the Vapnik-Chervonenkis (VC) dimension \cite{vapnik1999nature} or raw parameter counts tend to overestimate capacity in overparameterized neural networks. Zhang et al. \cite{zhang2017understanding} show that for over-parameterized deep networks—such as an Inception-style model with over a million parameters—straightforward VC-based bounds can vastly exceed their observed generalization performance. Part of this gap reflects how many directions in the networks’ high-dimensional weight space are effectively ‘flat,’ so moving along them does not meaningfully change the model’s outputs. Consequently, although these architectures can memorize random labels (yielding very large capacity measures), they still generalize well on real data, undermining naive uniform-capacity explanations

To better capture local sensitivity, curvature-aware approaches leverage the Fisher Information Matrix (FIM) \cite{amari1998natural}. We use the Two-Scale Effective Dimension (2SED), a  metric designed to complement existing measures (e.g., the Hessian trace \cite{liang2019fisher}) by integrating global parameter counts with local curvature effects encoded in the FIM. Unlike K-FAC \cite{martens2015optimizing}, which approximates the FIM for gradient-based updates, 2SED focuses on model complexity.

In modern deep learning, classical measures of model complexity (e.g., naive parameter counts or the VC dimension \cite{vapnik1999nature}) often fail to capture the non-uniform geometry and overparameterized nature of large-scale neural networks. Many directions in high-dimensional parameter spaces have negligible influence on the model’s behavior, while a comparatively small subset of ``sensitive'' directions carry most of the learning signal. This discrepancy motivates curvature-aware approaches, based on the Fisher information matrix (FIM), which measures local sensitivity of model parameters. In this section, we use the 2SED as a way to reconcile global dimension counts with localized curvature effects. We then discuss how 2SED can be used to adapt the fractional order parameter in FOSGD, providing a more stable and geometry-aware optimization algorithm.

\subsection{Foundational Definitions}

\begin{definition}[Fisher Information \cite{datres2024two}]
\label{def:fisher-info}
For a statistical model \( p_{\vartheta}(x, y) \) with parameters \( \vartheta \in \Theta \subseteq \mathbb{R}^d \), assuming \( p_{\vartheta} \) is differentiable and non-degenerate, define the log-likelihood as
\[
\ell_{\vartheta}(x, y) = \log p_{\vartheta}(x, y).
\]
Then, the \emph{Fisher Information Matrix} \( F(\vartheta) \) is given by
\begin{equation}
\label{eq:FisherInfo}
F(\vartheta) \;=\; \mathbb{E}_{(x, y) \sim p_{\vartheta}} 
\Bigl[ 
\bigl( \nabla_{\vartheta} \ell_{\vartheta}(x, y) \bigr) 
\otimes 
\bigl( \nabla_{\vartheta} \ell_{\vartheta}(x, y) \bigr) 
\Bigr],
\end{equation}
where \( \otimes \) denotes the outer product and the expectation is over \( p_{\vartheta} \). Under standard regularity conditions, this is equivalent to \( \mathbb{E}[-\nabla_{\vartheta}^2 \ell_{\vartheta}(x, y)] \) \cite{amari1998natural}.
\end{definition}

\begin{definition}[Empirical Fisher \cite{datres2024two}]
\label{def:empirical-fisher}
Given an i.i.d.\ sample \( \{(X_i, Y_i)\}_{i=1}^N \), the \emph{empirical Fisher Information Matrix} is
\begin{equation}
\label{eq:EmpiricalFisher}
F_N(\vartheta) \;=\; \frac{1}{N} \sum_{i=1}^N 
\Bigl( \nabla_{\vartheta} \ell_{\vartheta}(X_i, Y_i) \Bigr) 
\;\otimes\;
\Bigl( \nabla_{\vartheta} \ell_{\vartheta}(X_i, Y_i) \Bigr),
\end{equation}
which converges to the population Fisher in Eq.~\eqref{eq:FisherInfo} as \( N \to \infty \) (by the law of large numbers).
\end{definition}

\begin{definition}[Normalized Fisher Matrix \cite{datres2024two}]
\label{def:normalized-Fisher}
The \emph{normalized Fisher matrix} \( \widehat{F}(\vartheta) \) rescales \( F(\vartheta) \) so that
\[
\mathbb{E}_{\vartheta}\bigl[\mathrm{Tr}\,\widehat{F}(\vartheta)\bigr] \;=\; d,
\]
where \(d = \dim(\Theta)\). Formally,
\begin{equation}
\label{eq:NormalizedF}
\widehat{F}(\vartheta) \;=\;
\begin{cases}
\dfrac{d}{\mathbb{E}_{\vartheta} [\mathrm{Tr} \, F(\vartheta)]}\,F(\vartheta), 
& \text{if } \mathbb{E}_{\vartheta}[\mathrm{Tr}\,F(\vartheta)] > 0,\\[6pt]
0, 
& \text{otherwise}.
\end{cases}
\end{equation}
In practice, one approximates the expectation via Monte Carlo sampling (e.g., mini-batches).
\end{definition}
\subsection{The 2SED Approach}
\label{subsec:2sed}

Although \( d = \dim(\Theta) \) measures the \emph{nominal} number of parameters, many directions in parameter space may be relatively flat \cite{karakida2019universal} and do not strongly influence the loss. Consequently, 2SED integrates a curvature-based term derived from the Fisher matrix with the standard count \(d\). This approach captures the intuition that only certain directions in the parameter landscape are ``active'' in determining model capacity.

\begin{definition}[Two-Scale Effective Dimension \cite{datres2024two}]
\label{def:2sed}
Let \( \widehat{F}(\vartheta) \) be the normalized Fisher matrix, guaranteed positive semi-definite under mild conditions. For \( 0 \leq \zeta < 1 \), \( 0 < \varepsilon < 1 \), and \( \xi > 0 \), define the 2SED:
\begin{equation}
\label{eq:2SED}
d_{\zeta}(\varepsilon) \;=\; \zeta\,d \;+\; (1 - \zeta)\,d_{\text{curv}}(\varepsilon),
\end{equation}
where
\begin{equation}
\label{eq:dcurv}
d_{\text{curv}}(\varepsilon) 
\;=\; 
\frac{
  \log \mathbb{E}_{\vartheta} \Bigl[ 
    \det \Bigl( I_d + \varepsilon^{-\xi} \,\widehat{F}(\vartheta)^{\tfrac{1}{2}} \Bigr) 
  \Bigr]
}{
  \bigl|\log \bigl(\varepsilon^{\xi}\bigr)\bigr|
}.
\end{equation}
Here, \( I_d \) is the identity matrix, and \( \widehat{F}(\vartheta)^{\tfrac{1}{2}} \) is the PSD square root of \(\widehat{F}(\vartheta)\).
\end{definition}

\subsubsection*{Rationale for the Log-Determinant Term}
The use of \(\log \det\) reflects a \emph{compressive} summary of the spectrum of \(\widehat{F}(\vartheta)^{\tfrac{1}{2}}\). In particular, \(\log \det(I_d + A)\) can be seen as a measure of the ``total significant eigenvalues'' in \(A\). Directions with very small eigenvalues contribute minimally, whereas large-eigenvalue directions are emphasized. This aligns with information geometry principles, where \(\det(\mathrm{Id} + \mathrm{Fisher\ terms})\) captures local volume or effective degrees of freedom in parameter space \cite{amari1998natural}.

\subsubsection*{Derivation and Intuition}
Rewriting \(\det(I_d + A)\) as \(\prod_i \bigl(1 + \lambda_i(A)\bigr)\) (for diagonalizable matrices) yields:
\begin{equation}
\label{eq:dcurv_diag}
d_{\text{curv}}(\varepsilon) 
\;=\; 
\frac{
  \sum_{i=1}^d \log\bigl(1 + \varepsilon^{-\xi}\,\lambda_i^{1/2}\bigr)
}{
  \bigl|\log\bigl(\varepsilon^{\xi}\bigr)\bigr|
}.
\end{equation}
Hence, large \(\lambda_i\) (indicating high curvature directions) dominate the numerator. The scalar \(\zeta\) in Eq.~\eqref{eq:2SED} balances the nominal dimension \(d\) (e.g., the raw parameter count) and the Fisher-based curvature measure \(d_{\text{curv}}(\varepsilon)\). As \(\varepsilon \to 0\), curvature-based modes receive greater weighting. Meanwhile, if \(\zeta\) approaches 1, we recover a measure closer to \(d\) alone.

\section{The FOSGD and its modification}
\label{sec:frac_sgd_highlvl}

Classical stochastic gradient descent (SGD) updates parameters with the \emph{instantaneous} gradient. However, real-world optimization processes often exhibit memory effects, suggesting that a longer history of gradients might help or hinder convergence in non-trivial ways. \emph{Fractional calculus} supplies powerful operators, such as the Caputo derivative, that encode history in a mathematically principled manner. The fractional order \(\alpha\in(0,1)\) modulates how much past gradient information is used.

\subsection{Caputo Fractional Derivative and Fractional Updates}

\begin{definition}[Caputo Derivative \cite{monje2010fractional}]
\label{def:Caputo}
For \(n-1 < \alpha < n\), the \emph{Caputo fractional derivative} of a function \(f\) is\cite{Yang2023}
\begin{equation}
D_{t}^{\alpha}f(t)
\;=\;
\frac{1}{\Gamma(n - \alpha)}
\int_{0}^{t}
\frac{f^{(n)}(\tau)}{(t-\tau)^{\alpha - n + 1}}
\,d\tau.
\end{equation}
In many engineering and physics contexts, the Caputo form is preferred because it handles initial conditions naturally and the derivative of a constant is zero.
\end{definition}
In a discrete optimization setting, one can replace the classical gradient \(\nabla f(\Theta_t)\) by a \emph{fractional} gradient \(D_{t}^\alpha f(\Theta_t)\). Consider the fractional order SGD as follows \cite{Yang2023}:

\begin{equation}
\label{eq:frac_gd_update}
\Theta_{t+1}
\;=\;
\Theta_t
\;-\;
\eta\,D_{t}^{\alpha}f(\Theta_t).
\end{equation}
For $\alpha \in (0,1), \;\delta>0$ using Taylor series we have \cite{Yang2023}:
\begin{equation}
\label{eq:frac_gd_update_expanded}
\begin{split}
\Theta_{t+2}
&=
\Theta_{t+1}
-
\mu_t \frac{\nabla f(\Theta_{t+1})}{\Gamma(2-\alpha)} \\
&\quad\times
\Bigl(\lvert \Theta_{t+1} - \Theta_t\rvert + \delta\Bigr)^{1-\alpha}.
\end{split}
\end{equation}

The small offset \(\delta>0\) ensures that updates do not stall when consecutive iterates are nearly identical.

\subsection{Adapting the Fractional Exponent via 2SED}
\label{sec:modified_frac_sgd_2sed}
A fixed fractional exponent $\alpha$ can lead to instability if the model's curvature changes dramatically during training. Intuitively, high curvature or high 2SED indicates directions of rapid change or ``sensitivity,'' so reducing $\alpha$ helps prevent overshooting in these sensitive directions. By contrast, low curvature or low 2SED corresponds to flatter regions, where raising $\alpha$ leverages additional memory, potentially speeding up convergence by incorporating longer-horizon information. Hence, we propose a 2SED-based FOSGD that dynamically adjusts \(\alpha\) using 2SED of each layer. Suppose we compute the 2SED, \(d_{\zeta}^{(\ell)}(\varepsilon)\), for layer \(\ell\) and let
\begin{equation}
\label{eq:alpha_update}
\alpha_t^{(\ell)}
\;=\;
\alpha_{0}
\;-\;
\beta \times
\frac{
d_{\zeta}^{(\ell)}(\varepsilon)\big\rvert_{t}
}{
d_{\max}
},
\end{equation}
where \(\alpha_0\) is a base fractional order, \(\beta>0\) is a tuning parameter and \(d_{\max}\) is the maximum observed 2SED among all layers. The fraction \(\frac{d_{\zeta}^{(\ell)}(\varepsilon)\big\rvert_{t}}{d_{\max}}\) scales the current 2SED to the range \([0,1]\).
\subsection{Algorithm: The 2SEDFOSGD Algorithm }

Algorithm~\ref{alg:2sed_frac_sgd} provides a concise summary of the \emph{2SED-guided fractional SGD} approach, showing how each update leverages both fractional derivatives and curvature-based dimension signals.

{\small 
\begin{algorithm}[!hb]
\caption{The 2SEDFOSGD algorithm }
\label{alg:2sed_frac_sgd}
\begin{algorithmic}[1]
\Require 
\(
\begin{aligned}[t]
&\theta^0\\
&\alpha_0,\;\beta,\;\delta>0,\\
&\mu_0,\\
&\zeta,\,\varepsilon,\,\xi,\\
&t_{\max}, \\
\end{aligned}
\)

\State \textbf{Initialize:} \(\theta^1 \leftarrow \theta^0 - \mu_0\,\nabla f(\theta^0)\)\quad (Classical SGD update for the first step)
\For{\(t = 1,\dots,t_{\max}-1\)}
  \State \(\textbf{Compute gradient:}\quad g(\theta^t) \,\leftarrow\, \nabla f(\theta^t)\)
  \State \(\textbf{Compute Fisher Matrix:}\quad F^{(\ell)}(\theta^t)\;\;\forall \ell \in \{1,\dots,L\}\)
  \State \(\textbf{Compute the fractional exponent $\alpha_t^{(\ell)}$:}\)
\begin{equation}
\begin{split}
     \alpha_t^{(\ell)}
     &\;=\;
     \alpha_0
     \;-\;
     \beta
     \times
     \frac{ d_{\zeta}^{(\ell)}(\varepsilon)\big\rvert_{t} }{ d_{\max} }, \\
     d_{\max} &=
     \max_{\ell,\,k}\,d_{\zeta}^{(\ell)}(\varepsilon)\big\rvert_{k}.
\end{split}
\end{equation}

  \For{\(\ell = 1,\dots,L\)}
    \State\begin{equation}
\begin{split}
    \theta_{t+1}^{(\ell)}
    &\;=\;
    \theta_{t}^{(\ell)}
    \;-\;
    \frac{\mu_t}{\Gamma\bigl(2 - \alpha_t^{(\ell)}\bigr)} \\
    &\quad\times
    \bigl(
      \bigl\lvert \theta_t^{(\ell)} - \theta_{t-1}^{(\ell)} \bigr\rvert 
      + \delta
    \bigr)^{\,1 - \alpha_t^{(\ell)}}
    \,g^{(\ell)}(\theta^t).
\end{split}
\end{equation}

  \EndFor
\EndFor
\State \textbf{Output:}\quad \(\theta^{(t_{\max})}\)

\end{algorithmic}
\end{algorithm}

}

{\small
\section{Convergence Analysis for Non-Convex Objectives}
\label{sec:convergence_nonconvex}

We extend the analysis of \emph{Layerwise 2SED-Modified Fractional SGD} to non-convex objectives, proving convergence to critical points under realistic assumptions. This section is self-contained, with all proofs detailed inline for clarity.

\subsection{Foundational Definitions and Assumptions}

\paragraph{Non-Convex Objective.} Let \( f(\theta): \mathbb{R}^d \to \mathbb{R} \) be a non-convex, differentiable function, where \( \theta = (\theta^1, \dots, \theta^L) \), \( \theta^j \in \mathbb{R}^{d_j} \), and \( \sum_{j=1}^L d_j = d \). We assume:
\begin{itemize}
    \item \textbf{\( L \)-Smoothness:} For all \( \theta, \theta' \in \mathbb{R}^d \),
    \[
    \|\nabla f(\theta) - \nabla f(\theta')\| \le L \|\theta - \theta'\|,
    \]
    ensuring the gradient is Lipschitz continuous with constant \( L > 0 \).
    \item \textbf{Bounded Gradients:} For all \( \theta \in \mathbb{R}^d \),
    \[
    \|\nabla f(\theta)\| \le G,
    \]
    a common assumption in neural network training when gradients are clipped.
\end{itemize}

\paragraph{Stochastic Gradient Bounds.} The stochastic gradient estimator \( g(\theta^t) = (g^1(\theta^t), \dots, g^L(\theta^t)) \) satisfies:
\[
    \mathbb{E}[g^j(\theta^t)] = \nabla^j f(\theta^t), \quad \mathbb{E}[\|g^j(\theta^t) - \nabla^j f(\theta^t)\|^2] \le \sigma^2,
\]
for each layer \( j = 1, \dots, L \). Thus, the variance is bounded by \( \sigma^2 \), and:
\begin{equation}
\begin{split}
\mathbb{E}[\|g^j(\theta^t)\|^2] 
&= \mathbb{E}[\|g^j(\theta^t) - \nabla^j f(\theta^t) + \nabla^j f(\theta^t)\|^2] \\
&= \mathbb{E}[\|g^j(\theta^t) - \nabla^j f(\theta^t)\|^2] 
    \\
    &+ \|\nabla^j f(\theta^t)\|^2 \le \sigma^2 + G^2.
\end{split}
\end{equation}
implying \( \|g^j(\theta^t)\| \le \sqrt{\mathbb{E}[\|g^j(\theta^t)\|^2]} \le \sqrt{\sigma^2 + G^2} \).

\paragraph{Fractional Update Rule.} For layer \( j \), the update is:
\[
\theta^j_{t+1} = \theta^j_t - \eta_t^j g^j(\theta^t),
\]
where the effective step size is:
\[
\eta_t^j = \frac{\mu_t}{\Gamma(2-\alpha_j)} \bigl(\|\theta^j_t - \theta^j_{t-1}\| + \delta\bigr)^{1-\alpha_j},
\]
with \( \mu_t = \frac{\mu_0}{t^\rho} \), \( \delta > 0 \), and \( \alpha_j = \alpha_0 - \beta \frac{d^j_\zeta(\varepsilon)|_t}{d_{\max}} \), where \( \alpha_0 \in (0,1] \), \( d_{\max} = \max_{k,t} d^k_\zeta(\varepsilon)|_t \), and \( \beta > 0 \) ensures \( \alpha_j \in (0,1] \) (see Lemma~\ref{lemma:2sed_bound_nonconvex}). We bound:
\begin{itemize}
    \item \textbf{Gamma Function:} Since \( \alpha_j \in (0,1] \), \( 2-\alpha_j \in [1,2] \). The gamma function \( \Gamma(x) \) on \( [1,2] \) satisfies:
\begin{equation}
\begin{split}
c_\Gamma &= \Gamma(1) = 1, \\
C_\Gamma &= \max_{x \in [1,2]} \Gamma(x) \approx \Gamma(1.5) \\
&= \sqrt{\pi}/2 \approx 0.886 < 1.6.
\end{split}
\end{equation}
but conservatively, \( C_\Gamma = 1 \) at \( x = 2 \). Thus, \( 1 \le \Gamma(2-\alpha_j) \le 1.6 \).
    \item \textbf{Fractional Term:} Since \( \|\theta^j_t - \theta^j_{t-1}\| \le R_\Delta \) (Proposition~\ref{prop:bounded_iterates_nonconvex}), and \( 1-\alpha_j \in [0,1) \):
    \[
    \delta \le \|\theta^j_t - \theta^j_{t-1}\| + \delta \le \delta + R_\Delta,
    \]
    so:
    \[
    c_\Delta = \delta^{1-\alpha_{j,\max}}, \quad C_\Delta = (\delta + R_\Delta)^{1-\alpha_{j,\min}},
    \]
    where \( \alpha_{j,\max} = \max_{j,t} \alpha_j \), \( \alpha_{j,\min} = \min_{j,t} \alpha_j \), ensuring:
    \[
    c_\Delta \le \bigl(\|\theta^j_t - \theta^j_{t-1}\| + \delta\bigr)^{1-\alpha_j} \le C_\Delta.
    \]
\end{itemize}
Thus:
\[
\eta_t^j \in \left[ \mu_t \frac{c_\Delta}{C_\Gamma}, \mu_t \frac{C_\Delta}{c_\Gamma} \right].
\]

\paragraph{Step-Size Schedule.} Define \( \mu_t = \frac{\mu_0}{t^\rho} \), where \( 0.5 < \rho < 1 \), so:
\[
\sum_{t=1}^\infty \mu_t = \mu_0 \sum_{t=1}^\infty t^{-\rho} \ge \int_1^\infty x^{-\rho} dx = \frac{1}{\rho - 1} \to \infty,
\]
\begin{equation}
\begin{split}
\sum_{t=1}^\infty \mu_t^2 
&= \mu_0^2 \sum_{t=1}^\infty t^{-2\rho} \\
&\le \mu_0^2 \left( 1 + \int_1^\infty x^{-2\rho} dx \right) \\
&= \mu_0^2 \left( 1 + \frac{1}{2\rho - 1} \right) < \infty.
\end{split}
\end{equation}

since \( 2\rho > 1 \).

\subsection{Propositions and Lemmas}

\begin{proposition}[Bounded Iterates]
\label{prop:bounded_iterates_nonconvex}
For \( \mu_t = \frac{\mu_0}{t^\rho} \) (\( \mu_0 \) at \( t=0 \)), and \( \|g^j(\theta^t)\| \le \sqrt{\sigma^2 + G^2} \):
\[
\|\theta^j_t - \theta^j_{t-1}\| \le R_\Delta = \mu_0 \frac{C_\Delta}{c_\Gamma} \sqrt{\sigma^2 + G^2}.
\]
\begin{proof}
    For \( t = 1 \) (initial step):
    \[
    \theta^j_1 = \theta^j_0 - \mu_0 g^j(\theta^0),
    \]
    \[
    \|\theta^j_1 - \theta^j_0\| = \mu_0 \|g^j(\theta^0)\| \le \mu_0 \sqrt{\sigma^2 + G^2}.
    \]
    For \( t \ge 2 \):
    \[
    \|\theta^j_t - \theta^j_{t-1}\| = \eta_{t-1}^j \|g^j(\theta^{t-1})\|,
    \]
    where:
\begin{equation}
\begin{split}
\eta_{t-1}^j 
&= \frac{\mu_{t-1}}{\Gamma(2-\alpha_j)} 
    \bigl(\|\theta^j_{t-1} - \theta^j_{t-2}\| + \delta\bigr)^{1-\alpha_j} \\
&\le \mu_{t-1} \frac{C_\Delta}{c_\Gamma}.
\end{split}
\end{equation}

    since \( \Gamma(2-\alpha_j) \ge c_\Gamma = 1 \), and \( \bigl(\|\theta^j_{t-1} - \theta^j_{t-2}\| + \delta\bigr)^{1-\alpha_j} \le C_\Delta \). Now:
    \[
    \mu_{t-1} = \frac{\mu_0}{(t-1)^\rho}, \quad (t-1)^\rho \ge 1 \text{ for } t \ge 2,
    \]
    so \( \mu_{t-1} \le \mu_0 \). Thus:
    \[
    \eta_{t-1}^j \le \mu_0 \frac{C_\Delta}{c_\Gamma},
    \]
    \[
    \|\theta^j_t - \theta^j_{t-1}\| \le \mu_0 \frac{C_\Delta}{c_\Gamma} \sqrt{\sigma^2 + G^2} = R_\Delta.
    \]
    This holds for all \( t \ge 1 \), defining \( R_\Delta \) consistently.
\end{proof}
\end{proposition}

\begin{lemma}[Bounding the 2SED Measure]
\label{lemma:2sed_bound_nonconvex}
Under \( \mathbb{E}[\|g^j(\theta^t)\|^2] \le G^2 + \sigma^2 \), there exists a finite \( d_{\text{max,finite}} > 0 \) such that:
\[
d^j_\zeta(\varepsilon)|_t \le d_{\text{max,finite}}, \quad \forall t, j.
\]
\begin{proof}
    The 2SED is:
\begin{equation}
\begin{split}
d^j_\zeta(\varepsilon)|_t 
&= \zeta d_j + (1-\zeta) \\
    &\frac{\log \mathbb{E}_\theta [\det(I_{d_j} 
    + \varepsilon^{-\xi} \widehat{F}_j(\theta)^{1/2})]}
    {|\log(\varepsilon^\xi)|}.
\end{split}
\end{equation}

    where \( \widehat{F}_j(\theta^t) \approx \sum_{s=0}^{t-1} (1-\gamma)^s \gamma g^j(\theta^{t-s}) g^j(\theta^{t-s})^\top \), \( \gamma \in (0,1) \). The eigenvalues of \( g^j(\theta^{t-s}) g^j(\theta^{t-s})^\top \) are \( \|g^j(\theta^{t-s})\|^2 \) and 0s, so:
    \[
    \mathbb{E}[\|g^j(\theta^{t-s})\|^2] \le G^2 + \sigma^2.
    \]
    The weighted sum satisfies:
\begin{equation}
\begin{split}
\sum_{s=0}^{t-1} (1-\gamma)^s \gamma 
&= \gamma \sum_{s=0}^{t-1} (1-\gamma)^s \\
&= \gamma \frac{1 - (1-\gamma)^t}{1 - (1-\gamma)} \\
&= 1 - (1-\gamma)^t \le 1.
\end{split}
\end{equation}

    so the eigenvalues of \( \widehat{F}_j(\theta^t) \) are bounded by \( G^2 + \sigma^2 \). To bound the determinant, consider \( \det(I_{d_j} + \varepsilon^{-\xi} \widehat{F}_j(\theta)^{1/2}) \), where \( I_{d_j} \) is the \( d_j \times d_j \) identity matrix, \( \varepsilon^{-\xi} > 0 \) since \( \varepsilon \in (0,1) \) and \( \xi > 0 \), and \( \widehat{F}_j(\theta)^{1/2} \) is the positive semi-definite square root of the Fisher matrix \( \widehat{F}_j(\theta) \). The determinant of a matrix \( A \) is the product of its eigenvalues, so:
\[
\det(I_{d_j} + \varepsilon^{-\xi} \widehat{F}_j(\theta)^{1/2}) = \prod_{i=1}^{d_j} (1 + \varepsilon^{-\xi} \lambda_i^{1/2}),
\]
where \( \lambda_i \) are the eigenvalues of \( \widehat{F}_j(\theta) \), and thus \( \lambda_i^{1/2} \) are the eigenvalues of \( \widehat{F}_j(\theta)^{1/2} \), all non-negative because \( \widehat{F}_j(\theta) \) is positive semi-definite.

Next, we need an upper bound. From the previous step, the eigenvalues of \( \widehat{F}_j(\theta) \) are bounded by \( G^2 + \sigma^2 \), since \( \mathbb{E}[\widehat{F}_j(\theta)] \) has eigenvalues no larger than \( (G^2 + \sigma^2) [1 - (1-\gamma)^t] \le G^2 + \sigma^2 \). For a positive semi-definite matrix, the square root’s eigenvalues are the square roots of the original eigenvalues, so each \( \lambda_i \le G^2 + \sigma^2 \), and:
\[
\lambda_i^{1/2} \le \sqrt{G^2 + \sigma^2}.
\]
Since \( \varepsilon^{-\xi} > 0 \), the term \( 1 + \varepsilon^{-\xi} \lambda_i^{1/2} \) is increasing in \( \lambda_i^{1/2} \), and replacing each \( \lambda_i^{1/2} \) with its maximum possible value gives:
\[
1 + \varepsilon^{-\xi} \lambda_i^{1/2} \le 1 + \varepsilon^{-\xi} \sqrt{G^2 + \sigma^2}.
\]
Thus, the determinant is bounded by:
\begin{equation}
\begin{split}
\prod_{i=1}^{d_j} (1 + \varepsilon^{-\xi} \lambda_i^{1/2}) 
&\le \prod_{i=1}^{d_j} (1 + \varepsilon^{-\xi} \sqrt{G^2 + \sigma^2}) \\
&= (1 + \varepsilon^{-\xi} \sqrt{G^2 + \sigma^2})^{d_j}.
\end{split}
\end{equation}
because there are \( d_j \) eigenvalues, and the upper bound applies to each term in the product. Now, take the logarithm of the expected determinant. Since the determinant is positive (as \( 1 + \varepsilon^{-\xi} \lambda_i^{1/2} > 1 \)), and using Jensen’s inequality for the concave function \( \log \), we have \( \mathbb{E}[\log(\det(\cdot))] \le \log(\mathbb{E}[\det(\cdot)]) \), but here we directly bound the expectation of the determinant:
\begin{equation}
\begin{split}
\mathbb{E}[\det(I_{d_j} + \varepsilon^{-\xi} \widehat{F}_j(\theta)^{1/2})] 
&\le (1 + \varepsilon^{-\xi} \sqrt{G^2 + \sigma^2})^{d_j}.
\end{split}
\end{equation}

because the expectation preserves the bound derived from the eigenvalue estimate. Taking the logarithm:
\begin{equation}
\begin{split}
&\log \mathbb{E}[\det(I_{d_j} + \varepsilon^{-\xi} \widehat{F}_j(\theta)^{1/2})] \\
&\le \log \left[ (1 + \varepsilon^{-\xi} \sqrt{G^2 + \sigma^2})^{d_j} \right].
\end{split}
\end{equation}

Using the property of logarithms, \( \log(a^b) = b \log a \), this becomes:
\begin{equation}
\begin{split}
&\log \left[ (1 + \varepsilon^{-\xi} \sqrt{G^2 + \sigma^2})^{d_j} \right] \\
&= d_j \log(1 + \varepsilon^{-\xi} \sqrt{G^2 + \sigma^2}).
\end{split}
\end{equation}

so:
\[
\log \mathbb{E}[\det(\cdot)] \le d_j \log(1 + \varepsilon^{-\xi} \sqrt{G^2 + \sigma^2}).
\]

Finally, substitute this into the 2SED expression. The 2SED is:
\[
d^j_\zeta(\varepsilon)|_t = \zeta d_j + (1-\zeta) \frac{\log \mathbb{E}[\det(I_{d_j} + \varepsilon^{-\xi} \widehat{F}_j(\theta)^{1/2})]}{|\log(\varepsilon^\xi)|}.
\]
Since \( \varepsilon \in (0,1) \), \( \varepsilon^\xi < 1 \), \( \log(\varepsilon^\xi) < 0 \), and \( |\log(\varepsilon^\xi)| = -\log(\varepsilon^\xi) = \xi |\log \varepsilon| \) (noting \( \log \varepsilon < 0 \), so \( |\log \varepsilon| = -\log \varepsilon \)). Using the bound:
\[
d^j_\zeta(\varepsilon)|_t \le \zeta d_j + (1-\zeta) \frac{d_j \log(1 + \varepsilon^{-\xi} \sqrt{G^2 + \sigma^2})}{\xi |\log \varepsilon|},
\]
since \( \log \mathbb{E}[\det(\cdot)] \le d_j \log(1 + \varepsilon^{-\xi} \sqrt{G^2 + \sigma^2}) \), and the denominator is positive. Define:
\[
d_{\text{max,finite}} = \zeta d_j + (1-\zeta) \frac{d_j \log(1 + \varepsilon^{-\xi} \sqrt{G^2 + \sigma^2})}{\xi |\log \varepsilon|},
\]
which is a finite constant because \( \zeta \in [0,1) \), \( d_j \) is the layer dimension, \( \varepsilon \in (0,1) \), \( \xi > 0 \), and \( G^2 + \sigma^2 \) are fixed bounds, ensuring the expression is independent of \( t \) and \( j \).
\end{proof}
\end{lemma}

\begin{lemma}[Descent Lemma for Non-Convex Case]
\label{lemma:descent_nonconvex}
For \( L \)-smooth \( f \):
\begin{equation}
\begin{split}
\mathbb{E}[f(\theta^{t+1}) \mid \theta^t] 
&\le f(\theta^t) - \sum_{j=1}^L \eta_t^j \frac{c_\Gamma}{2 C_\Delta} \|\nabla^j f(\theta^t)\|^2 \\
&\quad + \sum_{j=1}^L (\eta_t^j)^2 \frac{C_\Delta^2}{c_\Gamma^2} (G^2 + \sigma^2).
\end{split}
\end{equation}

\begin{proof}
    By \( L \)-smoothness:
\begin{equation}
\begin{split}
f(\theta^{t+1}) 
&\le f(\theta^t) + \langle \nabla f(\theta^t), \theta^{t+1} - \theta^t \rangle \\
&\quad + \frac{L}{2} \|\theta^{t+1} - \theta^t\|^2.
\end{split}
\end{equation}

    where \( \theta^{t+1} - \theta^t = -\sum_{j=1}^L \eta_t^j g^j(\theta^t) \) (with appropriate vector stacking). Thus:
\begin{equation}
\begin{split}
f(\theta^{t+1}) 
&\le f(\theta^t) - \sum_{j=1}^L \eta_t^j \langle \nabla^j f(\theta^t), g^j(\theta^t) \rangle \\
&\quad + \frac{L}{2} \left\| \sum_{j=1}^L \eta_t^j g^j(\theta^t) \right\|^2.
\end{split}
\end{equation}

    Compute the squared norm:
\begin{equation}
\begin{split}
\left\| \sum_{j=1}^L \eta_t^j g^j(\theta^t) \right\|^2 
&= \sum_{j=1}^L (\eta_t^j)^2 \|g^j(\theta^t)\|^2 \\
&\quad + \sum_{i \neq j} \eta_t^i \eta_t^j \langle g^i(\theta^t), g^j(\theta^t) \rangle.
\end{split}
\end{equation}

    but for simplicity, bound it:
\begin{equation}
\begin{split}
\left\| \sum_{j=1}^L \eta_t^j g^j(\theta^t) \right\|^2 
&\le \left( \sum_{j=1}^L \eta_t^j \|g^j(\theta^t)\| \right)^2 \\
&\le L \sum_{j=1}^L (\eta_t^j)^2 \|g^j(\theta^t)\|^2.
\end{split}
\end{equation}

    by Cauchy-Schwarz. So:
\begin{equation}
\begin{split}
f(\theta^{t+1}) 
&\le f(\theta^t) - \sum_{j=1}^L \eta_t^j \langle \nabla^j f(\theta^t), g^j(\theta^t) \rangle \\
&\quad + \frac{L}{2} \sum_{j=1}^L (\eta_t^j)^2 \|g^j(\theta^t)\|^2.
\end{split}
\end{equation}

    Take expectations:
\begin{equation}
\begin{split}
\mathbb{E}[f(\theta^{t+1}) \mid \theta^t] 
&\le f(\theta^t) - \sum_{j=1}^L \eta_t^j \langle \nabla^j f(\theta^t), \nabla^j f(\theta^t) \rangle \\
&\quad + \frac{L^2}{2} \sum_{j=1}^L (\eta_t^j)^2 \mathbb{E}[\|g^j(\theta^t)\|^2].
\end{split}
\end{equation}

    since \( \mathbb{E}[g^j(\theta^t)] = \nabla^j f(\theta^t) \), and:
    \[
    \mathbb{E}[\|g^j(\theta^t)\|^2] \le G^2 + \sigma^2.
    \]
    Thus:
 \begin{equation}
\begin{split}
\mathbb{E}[f(\theta^{t+1}) \mid \theta^t] 
&\le f(\theta^t) - \sum_{j=1}^L \eta_t^j \|\nabla^j f(\theta^t)\|^2 \\
&\quad + \frac{L^2}{2} \sum_{j=1}^L (\eta_t^j)^2 (G^2 + \sigma^2).
\end{split}
\end{equation}

    Ensure \( \eta_t^j \le \frac{1}{L} \) (e.g., \( \mu_0 \frac{C_\Delta}{c_\Gamma} \le 1/L \)):
    \[
    1 - \frac{L}{2} \eta_t^j \ge 1 - \frac{L}{2} \cdot \frac{1}{L} = \frac{1}{2},
    \]
    but adjust directly:
\begin{equation}
\begin{split}
\mathbb{E}[f(\theta^{t+1}) \mid \theta^t] 
&\le f(\theta^t) \\
&-\sum_{j=1}^L \eta_t^j \left(1 - \frac{L}{2} \eta_t^j\right) \|\nabla^j f(\theta^t)\|^2 \\
& + \frac{L}{2} \sum_{j=1}^L (\eta_t^j)^2 (G^2 + \sigma^2).
\end{split}
\end{equation}

    Since \( 1 - \frac{L}{2} \eta_t^j \ge \frac{1}{2} \), and using bounds:
    \[
    \eta_t^j \ge \mu_t \frac{c_\Delta}{C_\Gamma}, \quad \eta_t^j \le \mu_t \frac{C_\Delta}{c_\Gamma},
    \]
    define \( c_1 = \frac{c_\Gamma}{2 C_\Delta} \), \( c_2 = \frac{C_\Delta^2}{c_\Gamma^2} (G^2 + \sigma^2) \):
\begin{equation}
\begin{split}
\mathbb{E}[f(\theta^{t+1}) \mid \theta^t] 
&\le f(\theta^t) - c_1 \sum_{j=1}^L \eta_t^j \|\nabla^j f(\theta^t)\|^2 \\
&\quad + c_2 \sum_{j=1}^L (\eta_t^j)^2.
\end{split}
\end{equation}

\end{proof}
\end{lemma}

\subsection{Main Convergence Theorem}

\begin{theorem}[Convergence in Non-Convex Setting]
\label{thm:main_conv_nonconvex}
Under the above assumptions:
\[
\min_{1 \le s \le T} \mathbb{E}[\|\nabla f(\theta^s)\|^2] = \mathcal{O}(1/T^{1-\rho}), \quad 0.5 < \rho < 1.
\]
\begin{proof}
    From Lemma~\ref{lemma:descent_nonconvex}:
\begin{equation}
\begin{split}
\mathbb{E}[f(\theta^{t+1}) \mid \theta^t] 
&\le f(\theta^t) - c_1 \sum_{j=1}^L \eta_t^j \|\nabla^j f(\theta^t)\|^2 \\
&\quad + c_2 \sum_{j=1}^L (\eta_t^j)^2,
\end{split}
\end{equation}

    where \( c_1 = \frac{c_\Gamma}{2 C_\Delta} \), \( c_2 = \frac{C_\Delta^2}{c_\Gamma^2} (G^2 + \sigma^2) \). Taking total expectation:
 \begin{equation}
\begin{split}
\mathbb{E}[f(\theta^{t+1})] 
&= \mathbb{E}[\mathbb{E}[f(\theta^{t+1}) \mid \theta^t]] \\
&\le \mathbb{E}[f(\theta^t)] - c_1 \mathbb{E}\left[ \sum_{j=1}^L \eta_t^j \|\nabla^j f(\theta^t)\|^2 \right] \\
&\quad + c_2 \mathbb{E}\left[ \sum_{j=1}^L (\eta_t^j)^2 \right].
\end{split}
\end{equation}

    Sum from \( t = 1 \) to \( T \):
\begin{equation}
\begin{split}
\sum_{t=1}^T \left( \mathbb{E}[f(\theta^{t+1})] - \mathbb{E}[f(\theta^t)] \right) 
&\le -c_1 \times\\
&\sum_{t=1}^T \mathbb{E}\left[ \sum_{j=1}^L \eta_t^j \|\nabla^j f(\theta^t)\|^2 \right] \\
&\quad + c_2 \sum_{t=1}^T \mathbb{E}\left[ \sum_{j=1}^L (\eta_t^j)^2 \right].
\end{split}
\end{equation}
So,
\begin{equation}
\begin{split}
\mathbb{E}[f(\theta^{T+1})] - \mathbb{E}[f(\theta^1)] 
&\le -c_1 \times\\
&\sum_{t=1}^T \mathbb{E}\left[ \sum_{j=1}^L \eta_t^j \|\nabla^j f(\theta^t)\|^2 \right] \\
&\quad + c_2 \sum_{t=1}^T \mathbb{E}\left[ \sum_{j=1}^L (\eta_t^j)^2 \right].
\end{split}
\end{equation}

    Since \( f \) is bounded below (\( f(\theta) \ge f_{\text{min}} \)):
\begin{equation}
\begin{split}
c_1 \sum_{t=1}^T \mathbb{E}\left[ \sum_{j=1}^L \eta_t^j \|\nabla^j f(\theta^t)\|^2 \right] 
&\le \mathbb{E}[f(\theta^1)] - \mathbb{E}[f(\theta^{T+1})] \\
&\quad + c_2 \sum_{t=1}^T \mathbb{E}\left[ \sum_{j=1}^L (\eta_t^j)^2 \right]\\
&\le f(\theta^1) - f_{\text{min}} \\
&+ c_2 L \sum_{t=1}^T (\eta_t^j)^2,
\end{split}
\end{equation}
    since \( \sum_{j=1}^L (\eta_t^j)^2 \le L \max_j (\eta_t^j)^2 \le L \mu_t^2 \frac{C_\Delta^2}{c_\Gamma^2} \). Compute:
    \[
    \sum_{t=1}^T \mu_t^2 = \mu_0^2 \sum_{t=1}^T t^{-2\rho} \le \mu_0^2 \left( 1 + \int_1^T x^{-2\rho} dx \right),
    \]
    \[
    \int_1^T x^{-2\rho} dx = \left[ \frac{x^{1-2\rho}}{1-2\rho} \right]_1^T = \frac{T^{1-2\rho} - 1}{1-2\rho},
    \]
    since \( 2\rho > 1 \), \( 1-2\rho < 0 \), and the integral is finite. Lower bound the weights:
    \[
    \sum_{t=1}^T \sum_{j=1}^L \eta_t^j \ge L \sum_{t=1}^T \mu_t \frac{c_\Delta}{C_\Gamma},
    \]
    and
\begin{equation}
\begin{split}
\sum_{t=1}^T \mu_t 
&= \mu_0 \sum_{t=1}^T t^{-\rho} \\
&\ge \mu_0 \int_1^T x^{-\rho} dx \\
&= \mu_0 \left[ \frac{x^{1-\rho}}{1-\rho} \right]_1^T \\
&= \mu_0 \frac{T^{1-\rho} - 1}{1-\rho}.
\end{split}
\end{equation}

    Thus:
\begin{equation}
\begin{split}
&\frac{1}{\sum_{t=1}^T \sum_{j=1}^L \eta_t^j} 
\sum_{t=1}^T \mathbb{E}\left[ \sum_{j=1}^L \eta_t^j \|\nabla^j f(\theta^t)\|^2 \right] \\
&\le \frac{f(\theta^1) - f_{\text{min}} 
+ c_2 L \mu_0^2 \frac{C_\Delta^2}{c_\Gamma^2} \frac{T^{1-2\rho} - 1}{1-2\rho}}
{c_1 L \mu_0 \frac{c_\Delta}{C_\Gamma} \frac{T^{1-\rho} - 1}{1-\rho}}.
\end{split}
\end{equation}

    Since \( 1-2\rho < 0 \), \( T^{1-2\rho} \to 0 \) as \( T \to \infty \), and:
\begin{equation}
\begin{split}
\min_{s=1,\dots,T} \mathbb{E}[\|\nabla f(\theta^s)\|^2] 
&\le \frac{\sum_{t=1}^T \mathbb{E}\left[ \sum_{j=1}^L \eta_t^j \|\nabla^j f(\theta^t)\|^2 \right]}
{\sum_{t=1}^T \sum_{j=1}^L \eta_t^j} \\
&= \mathcal{O}(1/T^{1-\rho}).
\end{split}
\end{equation}

\end{proof}
\end{theorem}
} 

\section{Experiments}
\label{sec:example1}

We evaluate the proposed algorithm performance in system identification for an AR(\(p\)) model, comparing results under both Gaussian and \(\alpha\)-stable (heavy-tailed) noise. The AR system under consideration is described by \cite{Yang2023}:
\begin{equation}
    y(k) = \sum_{i=1}^{p} a_i\,y(k - i) + \xi(k),
\end{equation}
where \(y(k - i)\) denotes the output at time \(k - i\), \(\xi(k)\) is a stochastic noise sequence, and \(a_i\) are unknown parameters. Our goal is to estimate these parameters. The corresponding regret (loss) function at time \(k\) is$J_k(\hat{\theta}) = \frac{1}{2}\,\bigl[y(k) - \phi^T(k)\,\hat{\theta}(k)\bigr]^2,$ where \(\hat{\theta}(k) = [\hat{a}_1(k), \ldots, \hat{a}_p(k)]^T\) and \(\phi(k) = [y(k - 1), \ldots, y(k - p)]^T\). In our experiments, we consider the specific second-order AR model$y(k) = 1.5\,y(k - 1) \;-\; 0.7\,y(k - 2) \;+\; \xi(k)$ so that the true coefficients are \(a_1 = 1.5\) and \(a_2 = -0.7\). We examine two noise distributions for \(\xi(k)\): Gaussian in Subsection~\ref{subsec:gaussian} and \(\alpha\)-stable (heavy-tailed) in Subsection~\ref{subsec:alpha_stable}. Throughout, we set \(\alpha_0 = 0.98\) and \(\beta = 0.01\) for our adaptive step-size parameters.

\subsection{Gaussian Noise}
\label{subsec:gaussian}

Under Gaussian white noise with zero mean and variance \(0.5\),
the proposed method converges reliably to the true coefficients,
highlighting its robustness and efficiency in standard stochastic settings.

\begin{figure}[!t]
    \centering
    \includegraphics[width=0.45\linewidth]{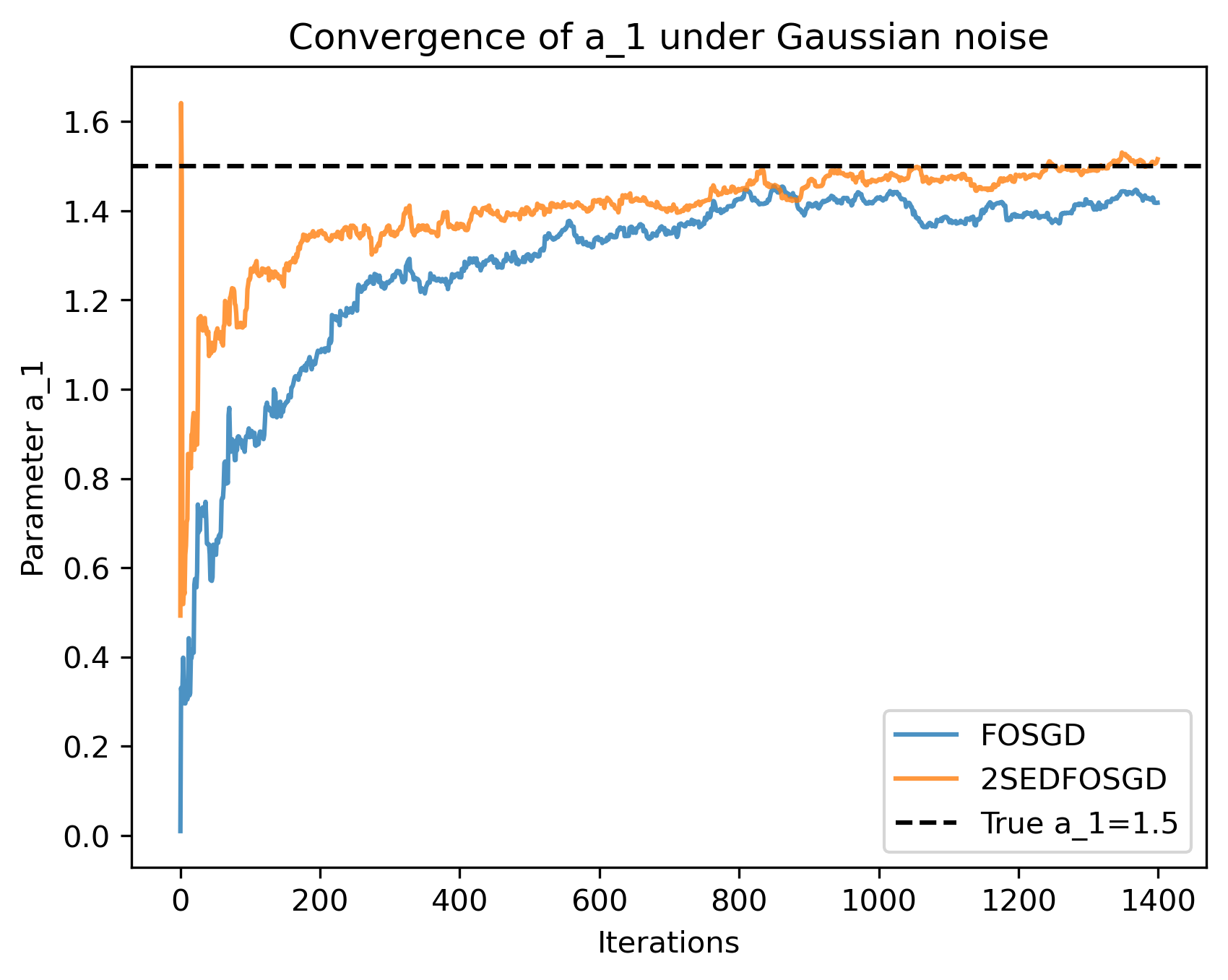}
    \hfill
    \includegraphics[width=0.45\linewidth]{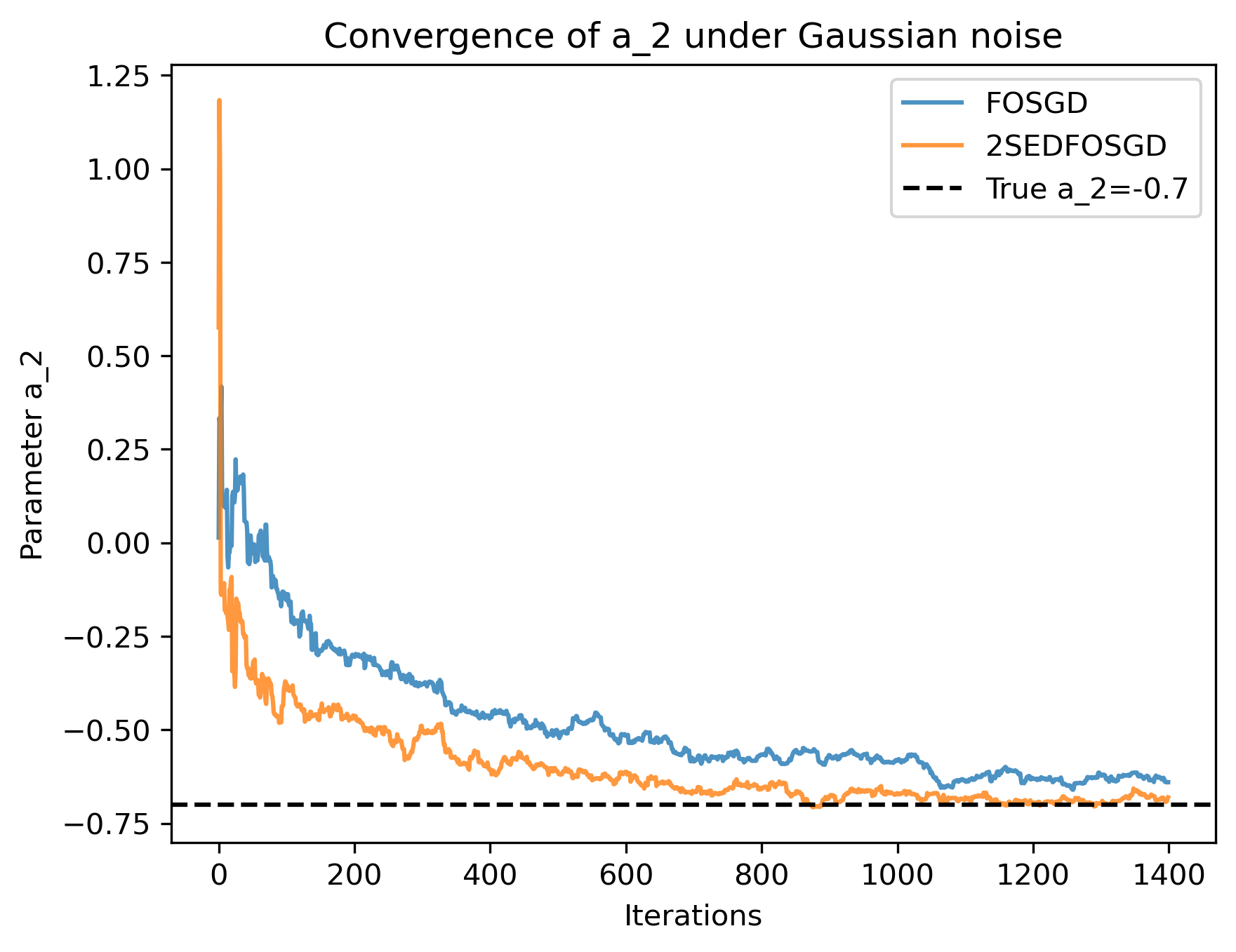}
    \caption{Under Gaussian noise at $\alpha=0.98$, both FOSGD and 2SEDFOSGD converge to the true AR parameters, with 2SEDFOSGD achieving faster and tighter estimates.}
    \label{fig:convex_ar2_gaussian_a1}
\end{figure}

\noindent
Figure~\ref{fig:convex_ar2_gaussian_a1} presents the convergence of FOSGD (blue) and 2SEDFOSGD (orange) toward the AR parameters $a_1$ and $a_2$ when $\alpha=0.98$ under Gaussian noise. While both algorithms ultimately reach accurate estimates, 2SEDFOSGD’s trajectories exhibit fewer transients and smoother adaptation.

\begin{figure}[!t]
    \centering
    \includegraphics[width=0.45\linewidth]{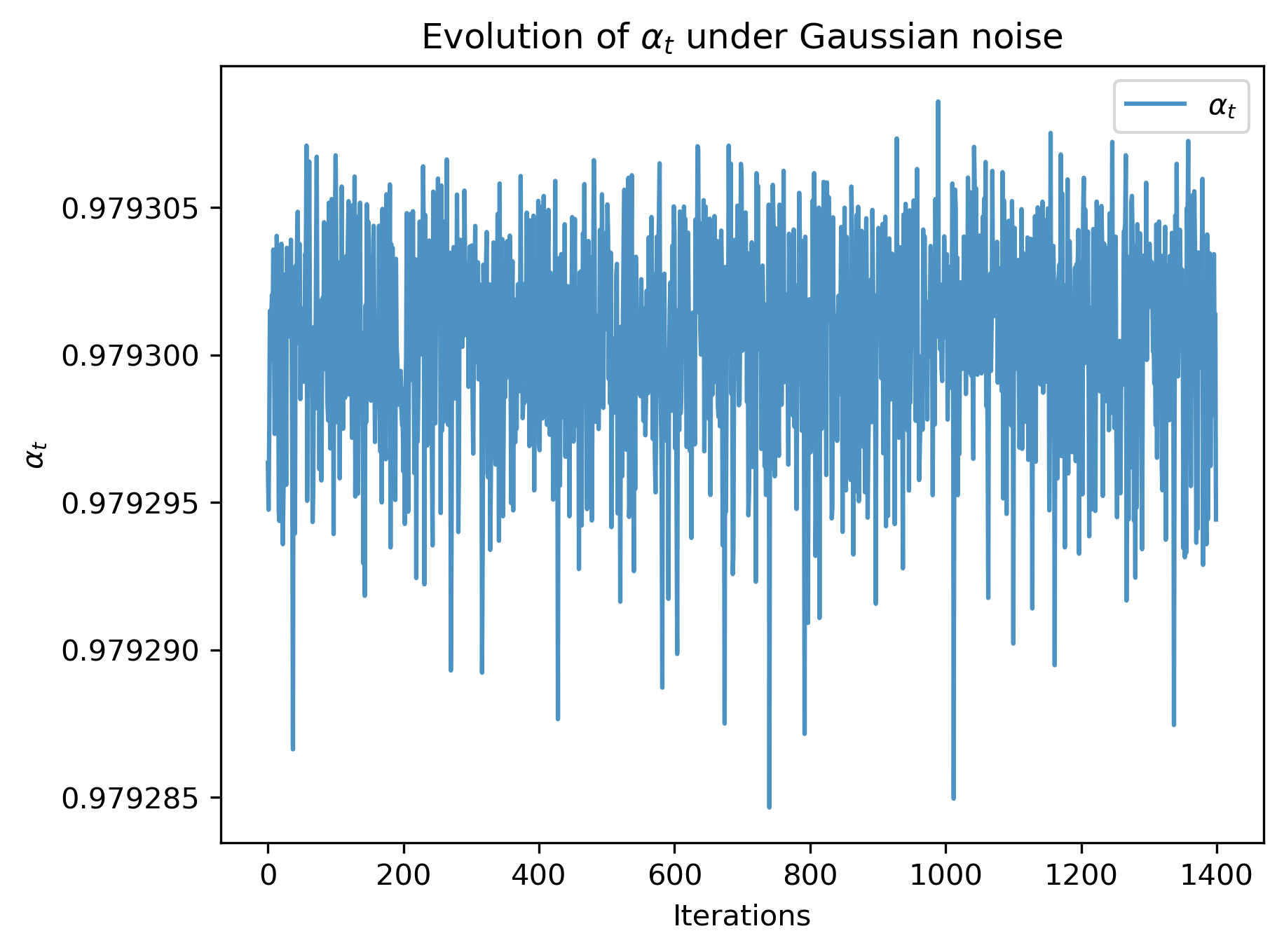}
    \hfill
    \includegraphics[width=0.45\linewidth]{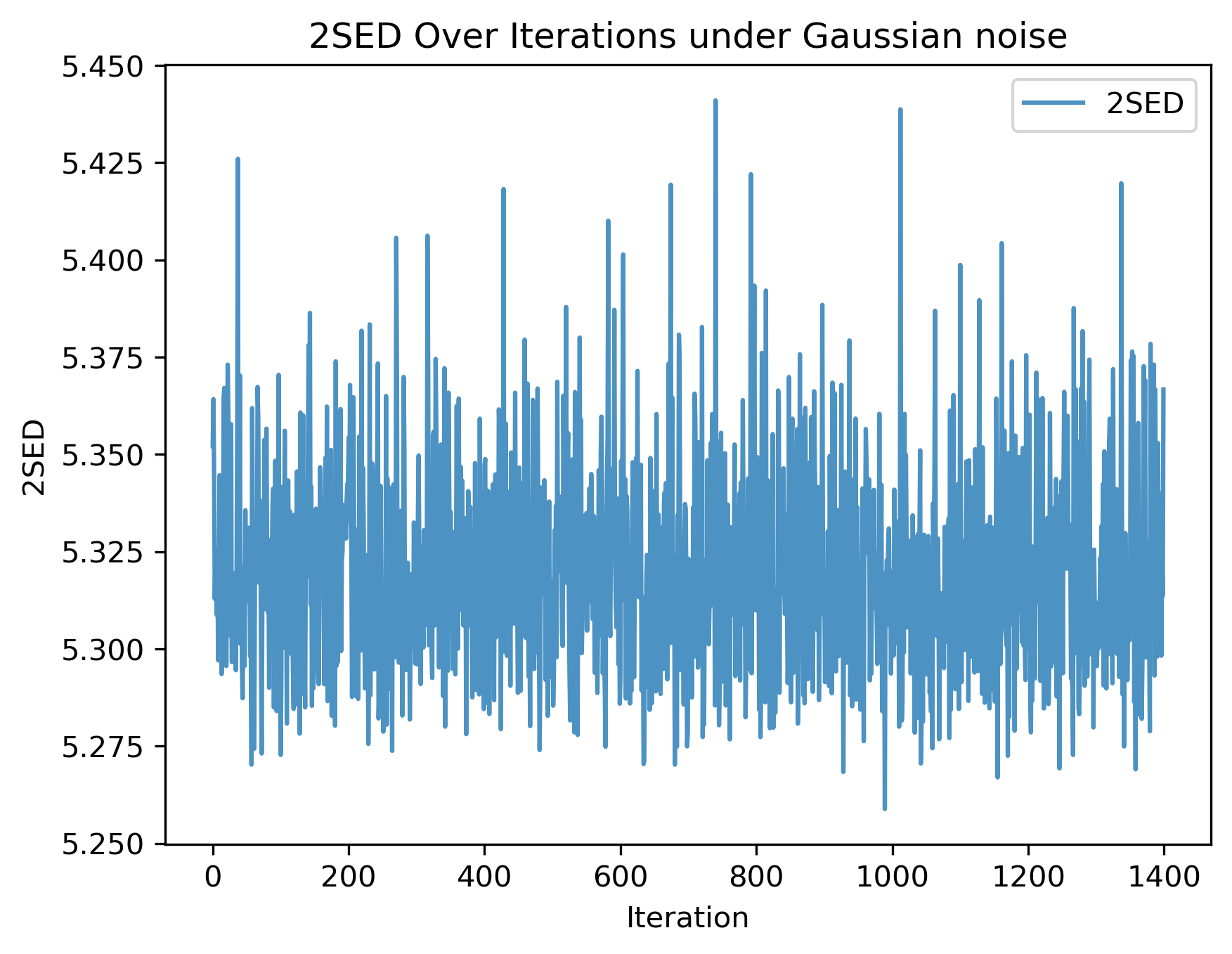}
    \caption{Effective Fractional Order $\alpha_t$ and 2SED Under Gaussian Noise ($\alpha=0.98$).
    Over 1{,}400 iterations, $\alpha_t$ remains near 0.9793, and the 2SED measure fluctuates within 5.25--5.45, indicating stable adaptive performance.}
    \label{fig:convex_ar2_gaussian_alpha_eff}
\end{figure}

\noindent
In Figure~\ref{fig:convex_ar2_gaussian_alpha_eff}, the adaptive fractional order $\alpha_t$ remains close to 0.9793, while the second-order extended dimension (2SED) measure oscillates between 5.25 and 5.45. This stability reflects balanced memory and curvature considerations over 1,400 iterations in a moderate-noise setting.

\begin{figure}[!t]
    \centering
    \includegraphics[width=0.45\linewidth]{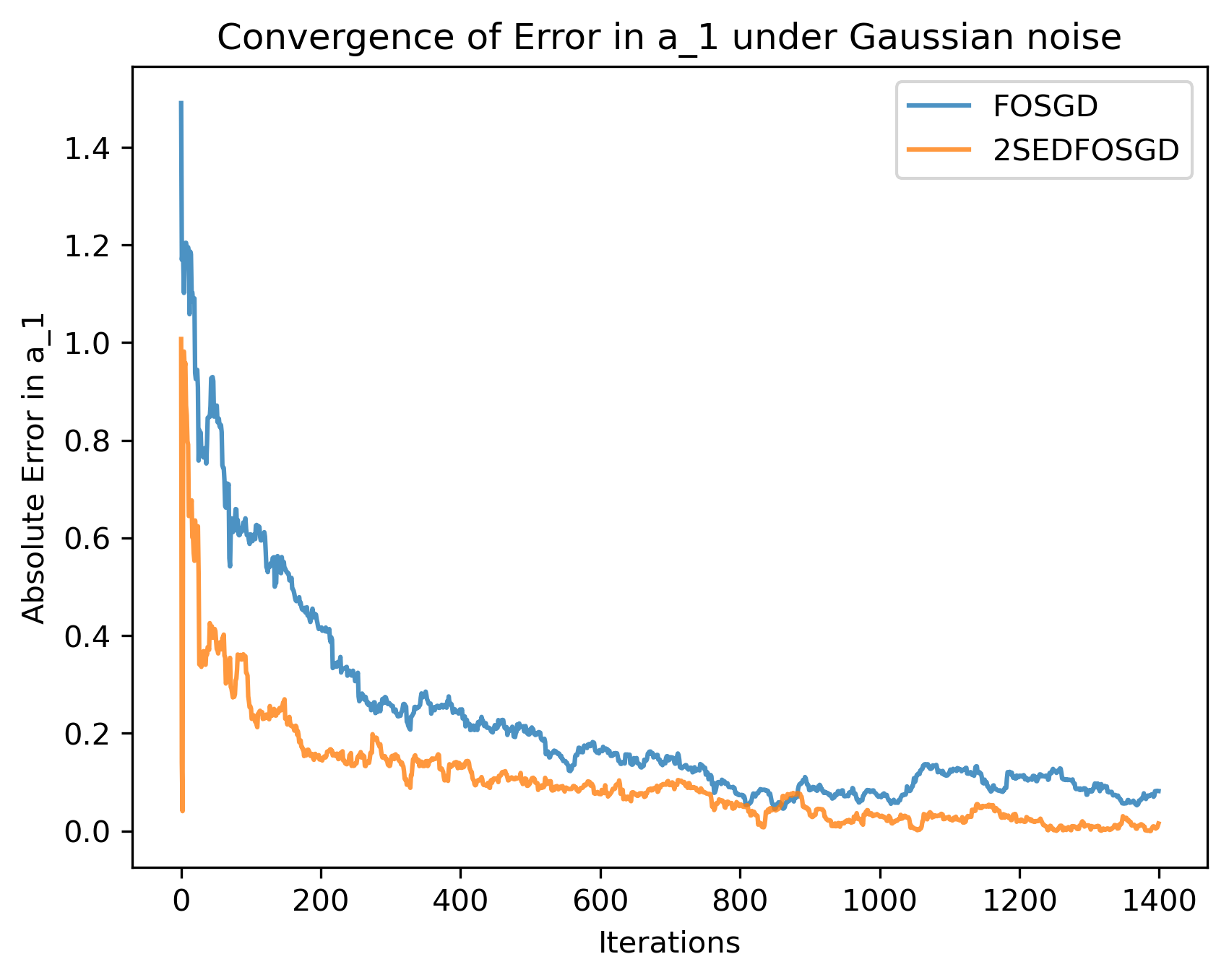}
    \hfill
    \includegraphics[width=0.45\linewidth]{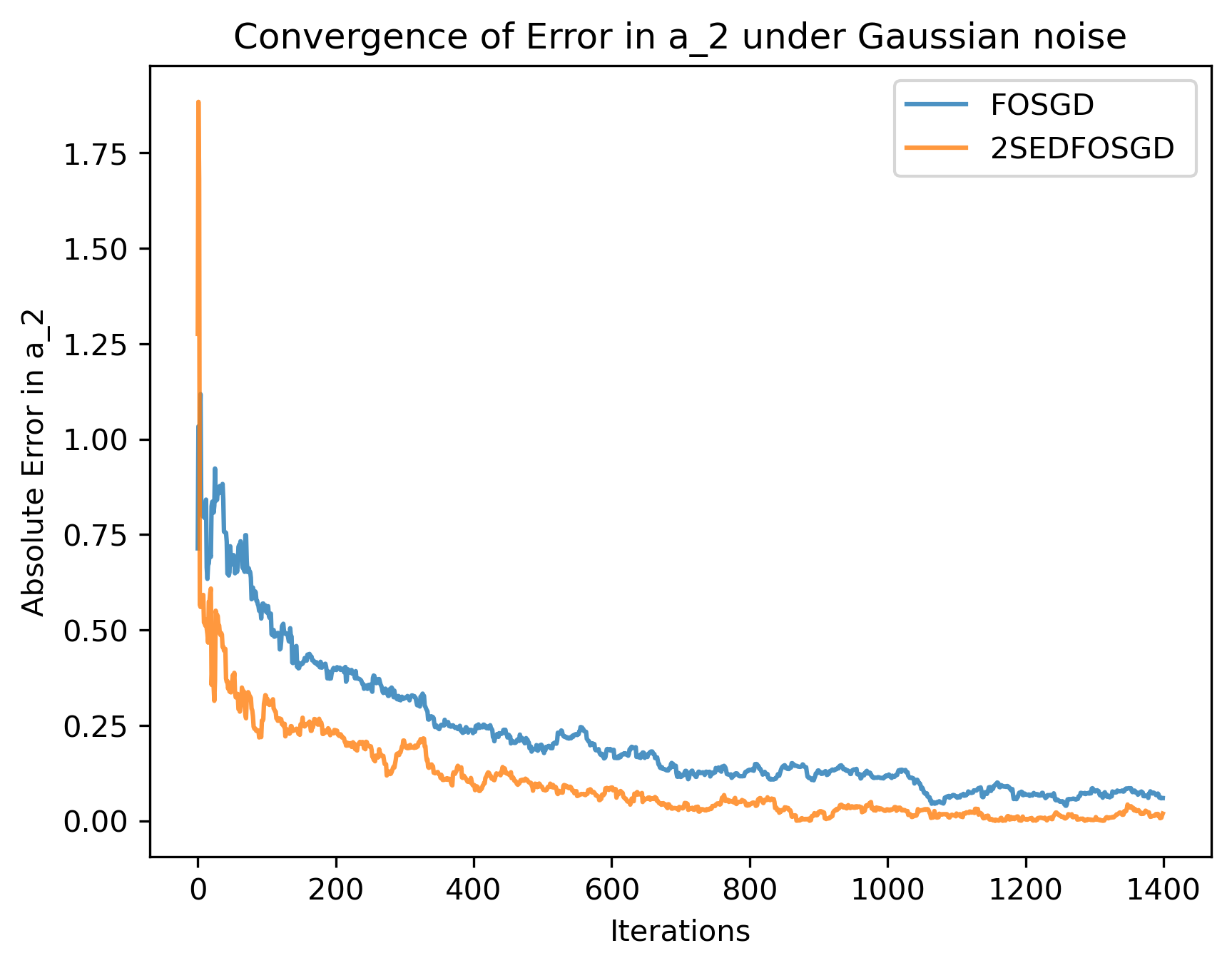}
    \caption{Under Gaussian noise at $\alpha=0.98$, 2SEDFOSGD maintains 
    lower absolute estimation errors than FOSGD for both AR parameters.}
    \label{fig:convex_ar2_gaussian_error_a1}
\end{figure}

\noindent
Figure~\ref{fig:convex_ar2_gaussian_error_a1} tracks the absolute errors in $a_1$ (left) and $a_2$ (right) across 1,400 iterations. Although both methods show initial spikes from random initialization, 2SEDFOSGD (orange) consistently remains closer to zero than FOSGD (blue). These findings highlight the benefit of incorporating second-order information to achieve more robust and accurate parameter estimates under Gaussian noise.

\subsubsection{Heavy-Tailed Noise} 
\label{subsec:alpha_stable}

Here, we extend our analysis to a convex setting where the noise follows a heavy-tailed distribution, which is common in adversarial environments or complex data scenarios. Our aim is to test the robustness of both FOSGD and 2SEDFOSGD under these challenging conditions.

\begin{figure}[!t]
    \centering
    \subfigure[$a_1$ Convergence]{
        \includegraphics[width=0.45\linewidth]{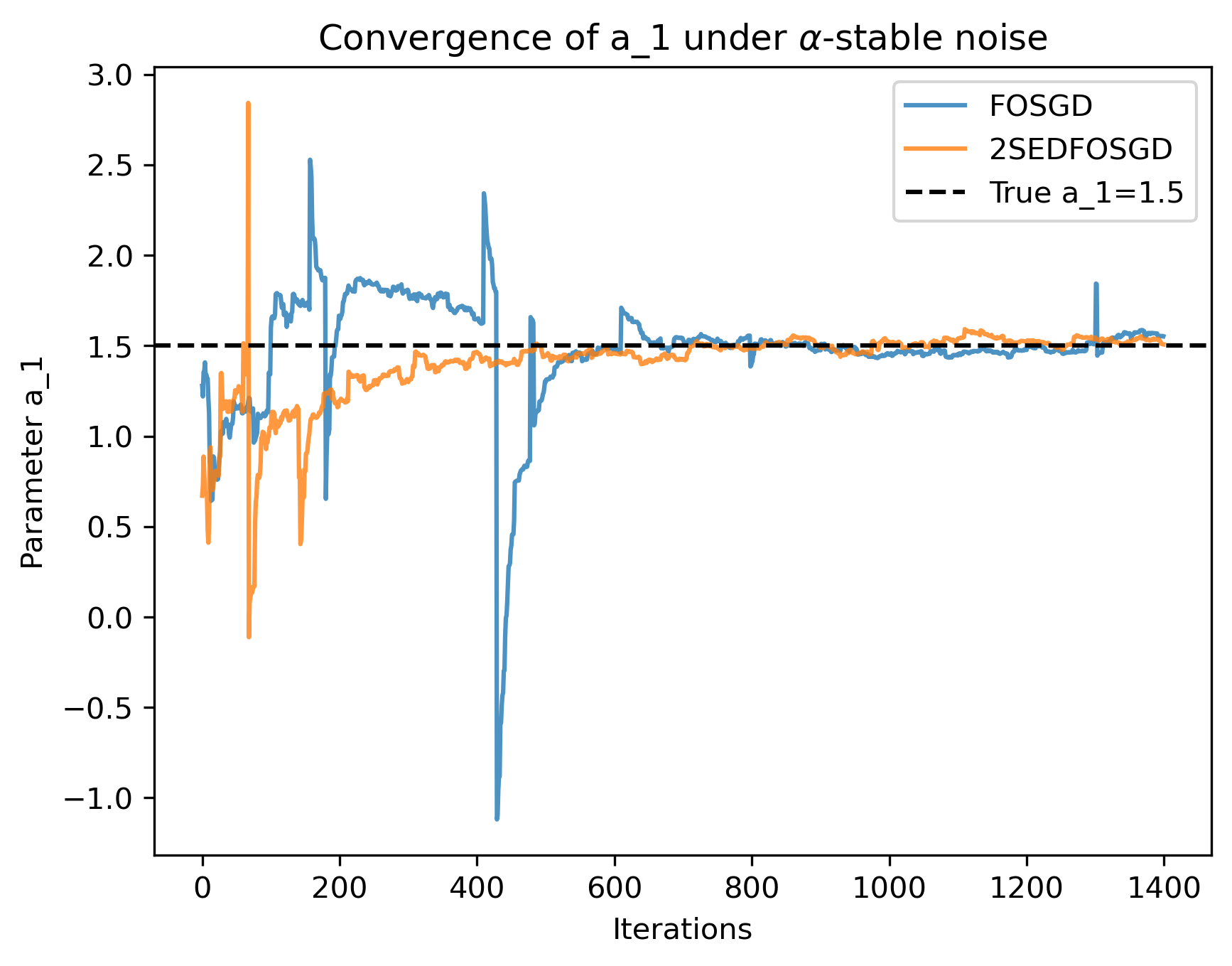}
    }
    \hfill
    \subfigure[$a_2$ Convergence]{
        \includegraphics[width=0.45\linewidth]{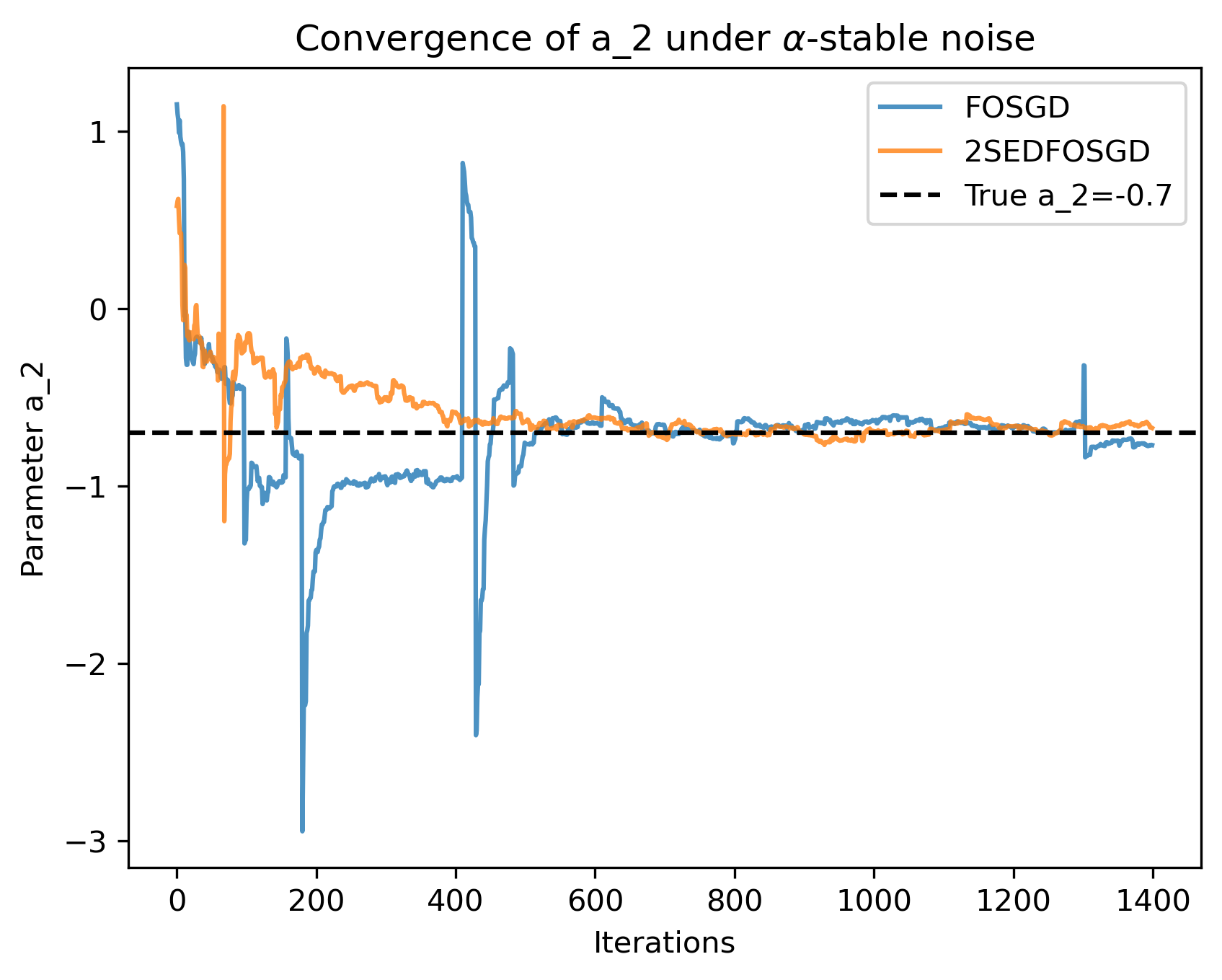}
    }
    \caption{Under $\alpha$-stable noise with $\alpha_{\mathrm{stbl}}=1.8$, 
    both FOSGD and 2SEDFOSGD converge near the true AR parameters, 
    with 2SEDFOSGD showing smoother trajectories and fewer extreme deviations.}
    \label{fig:ar2_a1_convex}
\end{figure}

Figure~\ref{fig:ar2_a1_convex} illustrates how FOSGD and 2SEDFOSGD handle $\alpha$-stable noise ($\alpha_{\text{stbl}}=1.8$). Although initial fluctuations occur, 2SEDFOSGD reaches stable estimates more quickly and remains closer to the true AR(2) parameters, highlighting its resilience under heavy-tailed disturbances.

\begin{figure}[!t]
    \centering
    \includegraphics[width=0.45\linewidth]{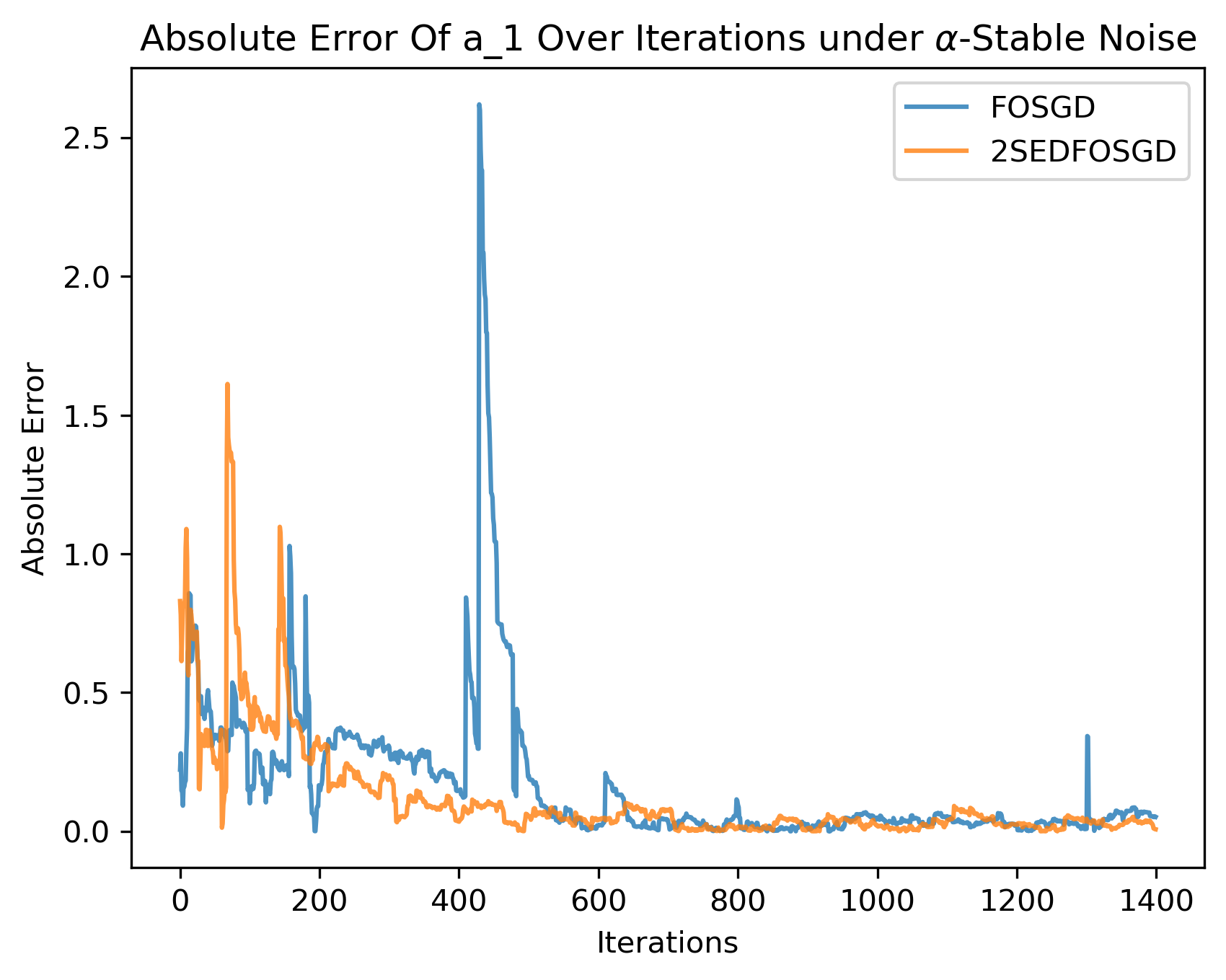}
    \hfill
    \includegraphics[width=0.45\linewidth]{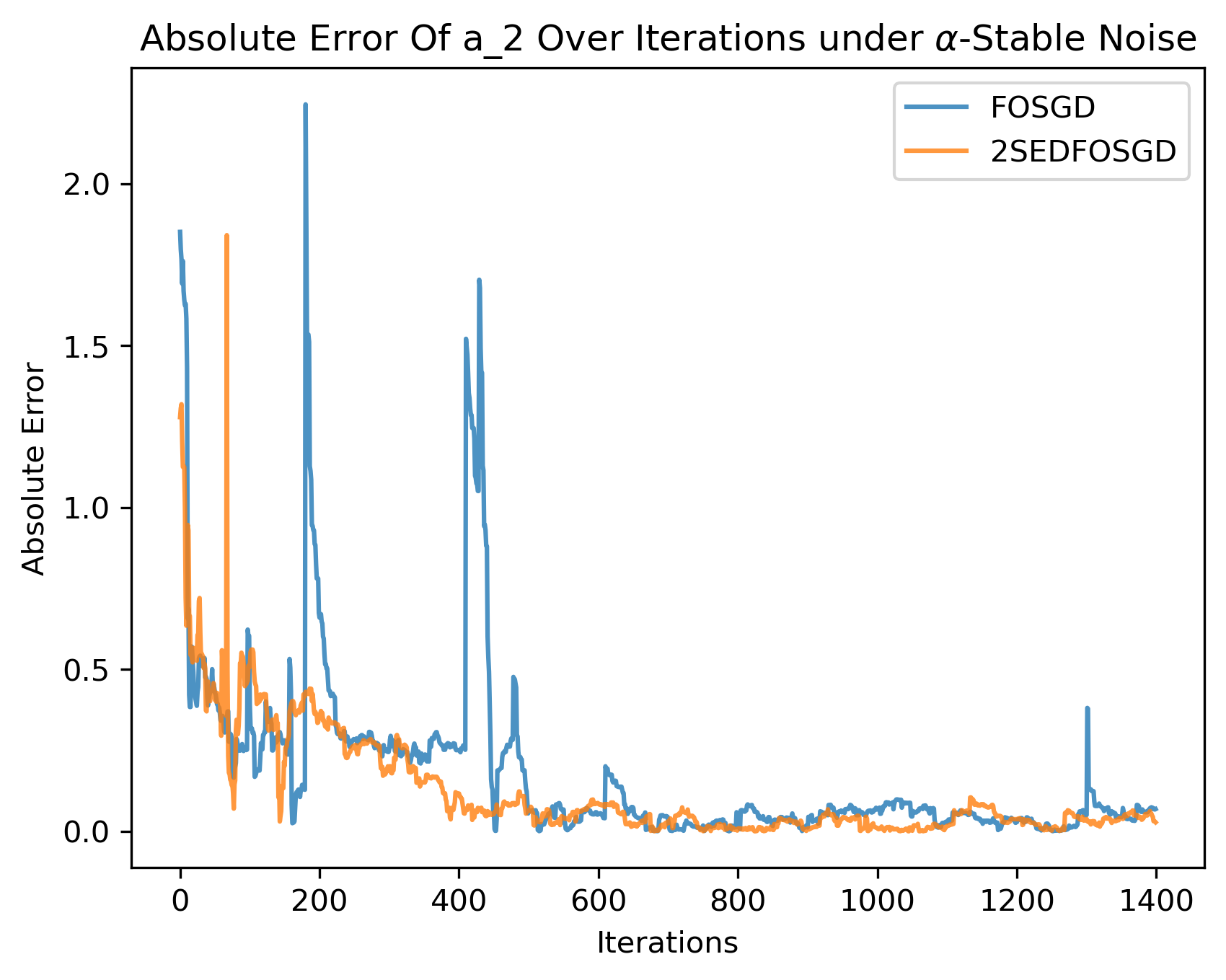}
    \caption{Under $\alpha_{\mathrm{stbl}}=1.8$, FOSGD shows higher oscillations before stabilizing, 
    while 2SEDFOSGD exhibits a smoother, consistently decreasing error trajectory for both 
    $a_1$ and $a_2$.}
    \label{fig:ar2_error_a1_convex}
\end{figure}

Figure~\ref{fig:ar2_error_a1_convex} compares the absolute estimation errors for $a_1$ (left) and $a_2$ (right). Although FOSGD eventually settles to a similar final error, 2SEDFOSGD displays a more gradual and steady decline, reflecting improved tolerance to large noise bursts.

\begin{figure}[!t]
    \centering
    \includegraphics[width=0.45\linewidth]{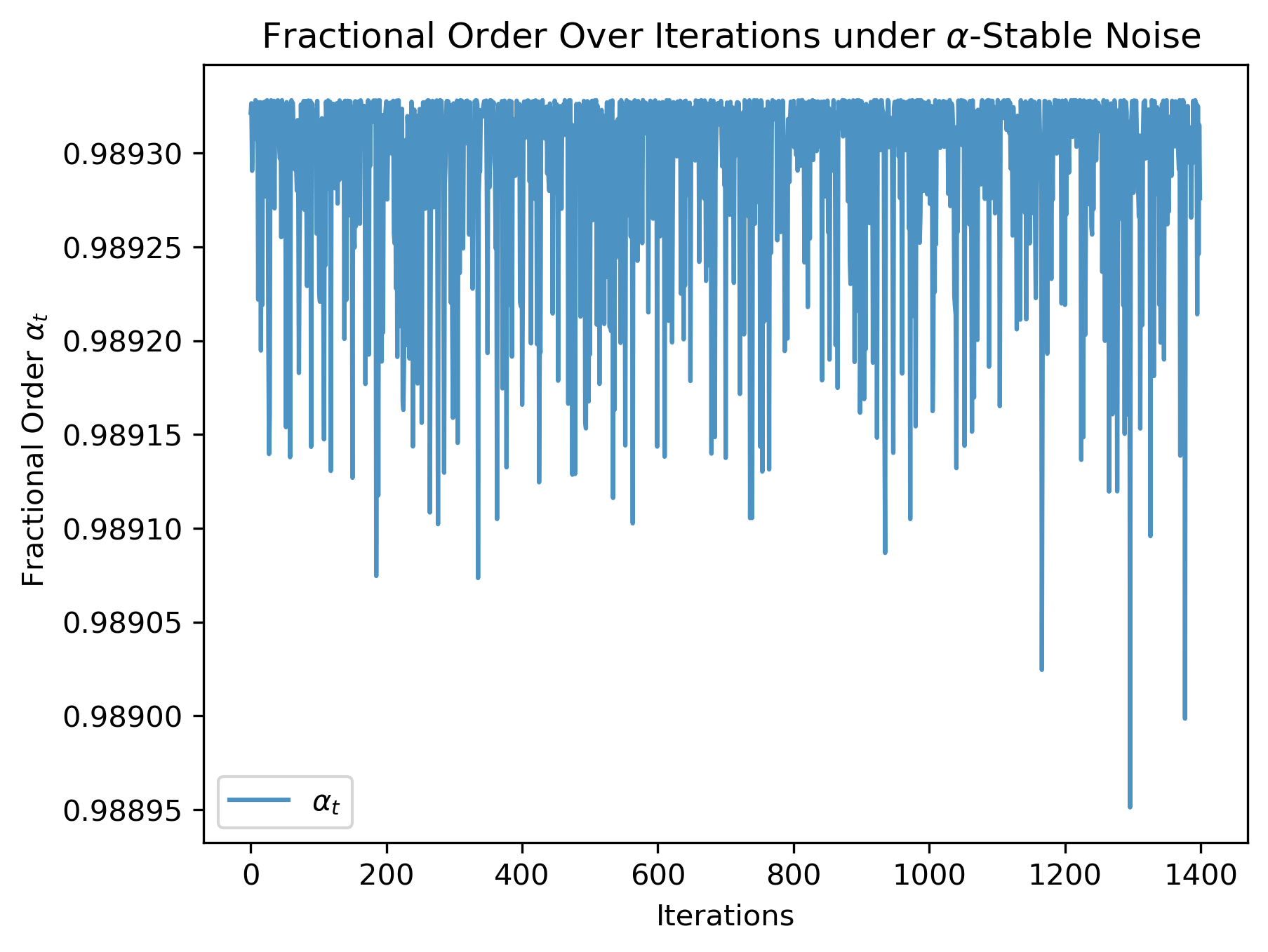}
    \hfill
    \includegraphics[width=0.45\linewidth]{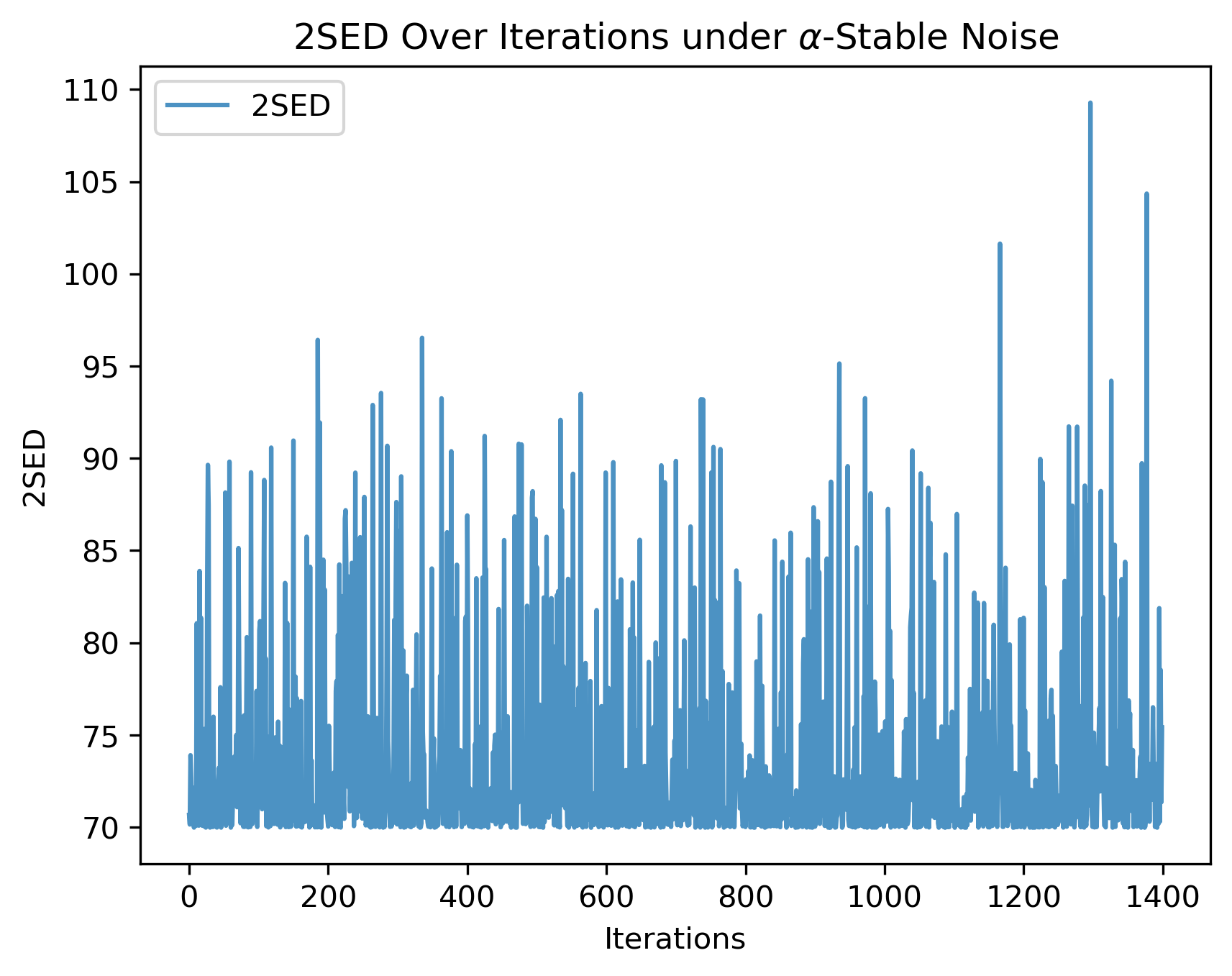}
    \caption{Under $\alpha$-stable noise ($\alpha_{\mathrm{stbl}}=1.8$), 2SEDFOSGD adapts 
    its internal parameters: $\alpha_t$ hovers near 0.989 (left), and the second-order 
    estimation dimension (2SED) evolves online (right).}
    \label{fig:ar2_alpha_eff_convex}
\end{figure}

Figure~\ref{fig:ar2_alpha_eff_convex} highlights 2SEDFOSGD’s internal adaptation under heavy-tailed conditions. The effective fractional order $\alpha_t$ remains close to 0.989, while the 2SED metric shifts in response to observed gradients, supporting robust performance.


\section{Conclusion and Future Work}
\label{sec:conclusion}
In this paper, we proposed the \textit{2SED Fractional-Order Stochastic Gradient Descent} (2SEDFOSGD) algorithm, 
which augments fractional-order SGD (FOSGD) with a Two-Scale Effective Dimension (2SED) framework to dynamically 
adapt the fractional exponent. By continuously monitoring model sensitivity and effective dimensionality, 2SEDFOSGD 
mitigates oscillatory or sluggish convergence behaviors commonly encountered with naive fractional approaches. 
We evaluated the performance of 2SEDFOSGD through a system identification task using an autoregressive (AR) model 
under both Gaussian and \(\alpha\)-stable noise. Our results demonstrated that the proposed 
dimension-aware fractional-order update scheme not only accelerates convergence but also achieves more robust 
parameter estimates than baseline methods in these noise scenarios.






\bibliographystyle{IEEEtran}
\bibliography{References} 
\end{document}